%% file: main.tex
\documentclass[conference]{IEEEtran}
\IEEEoverridecommandlockouts
\usepackage{lineno}
\usepackage{cite}
\usepackage{amsmath,amssymb,amsfonts}
\usepackage{algorithmic}
\usepackage{graphicx}
\usepackage{textcomp}
\usepackage{xcolor}
\usepackage{hyperref}
\usepackage{cleveref}
\usepackage{booktabs}
\usepackage{url}
\usepackage[square,numbers]{natbib}
\usepackage{caption}
\usepackage{graphicx}
\usepackage{subcaption}
\input{preamble}

\begin{document}

%%%%%%%%% TITLE - UPDATED FORMATTING
\title{\bfseries \LARGE REDistill: Robust Estimator Distillation for Balancing Robustness and Efficiency}

%%%%%%%%% AUTHORS - UPDATED FORMATTING
\author{Ondřej Týbl\\
Department of Cybernetics\\
FEE, Czech Technical University\\
{\tt\small tyblondr@cvut.cz}
% For a paper whose authors are all at the same institution,
% omit the following lines up until the closing ``}''.
% Additional authors and addresses can be added with ``\and'',
% just like the second author.
% To save space, use either the email address or home page, not both
\and
Lukáš Neumann\\
Department of Cybernetics\\
FEE, Czech Technical University\\
{\tt\small lukas.neumann@cvut.cz}
}

\maketitle

\input{sec/0_abstract}
\input{sec/1_introduction}
\input{sec/2_related_work}
\input{sec/3_background}
\input{sec/4_methodology}
\input{sec/5_experiments}
\input{sec/6_conclusion}
%
% WARNING: do not forget to delete the supplementary pages from your submission 

{\small
    \bibliographystyle{ieeenat_fullname}
    \bibliography{main}
}

\input{sec/X_suppl}

\end{document}

%% file: preamble.tex
\makeatletter
\renewcommand{\paragraph}{%
  \@startsection{paragraph}{4}%
  {\z@}{0.2em}{-1em}%
  {\normalfont\normalsize\bfseries}%
}
\makeatother

\usepackage{mathtools}

%% file: sec/0_abstract.tex
\begin{abstract}
Knowledge Distillation (KD) transfers knowledge from a large teacher model to a smaller student by aligning their predictive distributions. However, conventional KD formulations -- typically based on Kullback–Leibler divergence -- assume that the teacher provides reliable soft targets. In practice, teacher predictions are often noisy or overconfident, and existing correction-based approaches rely on ad-hoc heuristics and extensive hyper-parameter tuning, which hinders generalization.
We introduce \textbf{REDistill} (\textit{Robust Estimator Distillation}), a simple yet principled framework grounded in robust statistics. REDistill replaces the standard KD objective with a \emph{power divergence} loss, a generalization of KL divergence that adaptively downweights unreliable teacher output while preserving informative logit relationships. This formulation provides a unified and interpretable treatment of teacher noise, requires only logits, integrates seamlessly into existing KD pipelines, and incurs negligible computational overhead.
Extensive experiments on CIFAR-100 and ImageNet-1k demonstrate that REDistill consistently improves student accuracy in diverse teacher--student architectures. Remarkably, it achieves these gains \textbf{without model-specific hyper-parameter tuning}, underscoring its robustness and strong generalization to unseen teacher–student pairs.
\end{abstract}

%% file: sec/1_introduction.tex
\section{Introduction}
\label{sec:introduction}
% KD is important
Rapid development of deep neural networks has been coupled with a substantial increase in model size and therefore computational cost, posing challenges for efficient training and deployment of such models. Although larger models generally exhibit higher accuracy, considerable research effort has been devoted to reduce model complexity without sacrificing accuracy. Knowledge Distillation (KD) has recently gained significant research attention \cite{cao2023excellent, sun2024logit, sun2025knowledge, li2023curriculum}, because compared to other strategies such as pruning or quantization, knowledge distillation has fewer architectural constraints and therefore offers greater flexibility and wider applicability.

\input{figures/figure1}

% Teacher is noisy
In Knowledge Distillation, the knowledge accumulated in a pre-trained high-capacity \textit{teacher} model is transferred to a more efficient, smaller \textit{student} network. The main challenge, however, is that even the \textit{teacher's} predictions are not perfect, in other words, the \textbf{teacher is noisy}, and as a result, blindly mimicking teacher outputs only leads to error accumulation and to a significant degradation of the \textit{student} model accuracy.

% How was noisy teacher addressed so far
Existing correction-based distillation methods \cite{cao2023excellent, lan2025improve, wen2021preparing} adjust the teacher’s logits using ground-truth information to address this issue. Specifically, they either swap the logits between the predicted top class and the ground-truth class (the swap operation) \cite{wen2021preparing}, or amplify the probability assigned to the ground-truth class (the augment operation) \cite{cao2023excellent, lan2025improve} or more recently, \cite{sun2025knowledge} masks certain classes during distillation. However, these ad-hoc modifications and heuristics distort the underlying correlations among classes and their effectiveness is highly sensitive to hyper-parameter tuning, whose values need to be \textit{empirically determined for each teacher/student pair} and for \emph{each dataset}.

% How we address it
In this paper, we introduce a robust principled framework based on the power divergence family from robust statistics. By replacing the standard KD objective with its power divergence counterpart, we adaptively down-weight unreliable teacher outputs while preserving essential logit relationships. As a result, our method tackles mistakes of the noisy teacher network in a unified and an interpretable manner. Thanks to its fundamental grounding, the method is less sensitive to specific hyper-parameter values; even more so, we are able to determine the most important hyper-parameter value theoretically (and we empirically show that this choice is indeed optimal). Our method, without bells and whistles, not only outperforms previous KD methods, but it does so \textbf{without model-specific hyper-parameter tuning for each teacher/student pair}, which makes our method more universally applicable.

Our main contributions are summarized as follows:
\begin{itemize}
    \item We propose a new principled method for Knowledge Distillation, which compensates for unreliable teacher predictions. The method replaces the standard KD objective with its power divergence counterpart, which adaptively down-weights unreliable teacher outputs while preserving essential logit relationships.
    \item The method does not rely on model-specific hyper-parameter tuning and heuristics, which makes it more generalizable. Experiments in the model-agnostic setting suggest that the method will also likely work on unseen teacher/student pairs, because without a single modification and using hyper-parameter value which was first obtained theoretically, the method outperformed previous methods on 14 different teacher/student pairs.   % Kydz to vyslo dobre na vybrany modely, je sance ze to bude fungovat stejne dobre i na novy
\end{itemize}

The remainder of the paper is organized as follows. In~\Cref{sec:related_work}, we review related work on knowledge distillation and robust loss design.  \Cref{sec:background} introduces the necessary background and notation, and \Cref{sec:methodology} presents the proposed method in detail. Experimental results are provided in \Cref{sec:experiments}, followed by conclusions in \Cref{sec:conclusion}.

%% file: figures/figure1.tex
\begin{figure}
    \centering
    \centering
        \includegraphics[width=0.8\columnwidth]{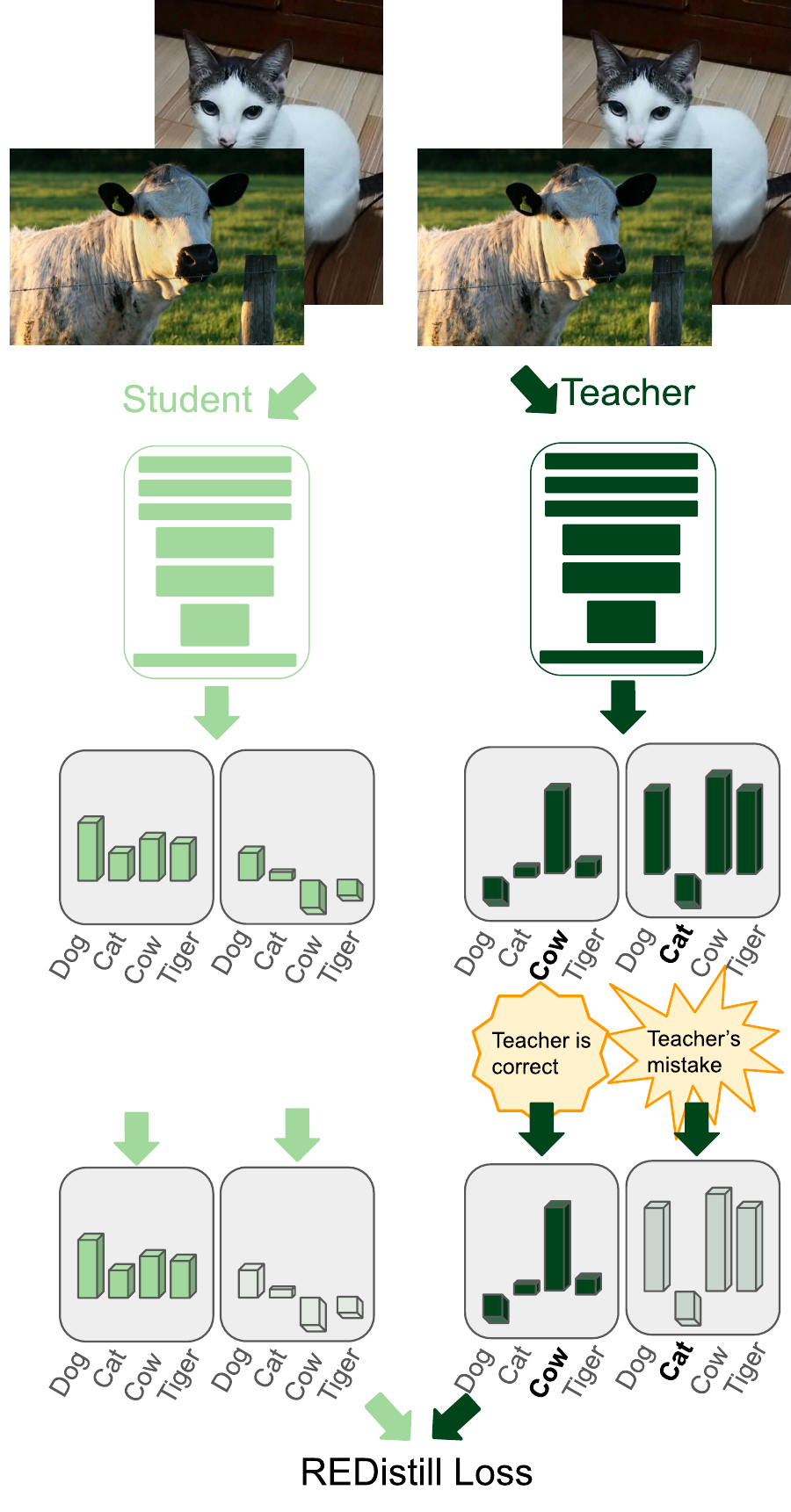}
    \caption{Knowledge distillation trains a smaller student model by matching its outputs to a larger teacher’s logits. However, teachers -- even large ones -- can be unreliable, leading to degraded student performance. Our method introduces a robust distillation loss that adapts the contribution of each training example based on how reliable the typical teacher’s prediction is for the given input. This enables the student to dynamically trust or distrust the teacher beyond simply checking whether the top logit is correct. Our approach is theoretically sound and simple to integrate with existing distillation methods.\vspace{-10pt}}
    \label{fig:figure1}
\end{figure}

%% file: sec/2_related_work.tex
\section{Related Work}
\label{sec:related_work}

Knowledge Distillation (KD) was first introduced by Hinton et al.~\cite{hinton2015distilling}, who proposed transferring knowledge from a high-capacity \textit{teacher} network to a smaller \textit{student} by minimizing the Kullback--Leibler (KL) divergence between their softened logit distributions. Zhao et al.~\cite{zhao2022decoupled} later refined this formulation by decoupling target and non-target logits, allowing the student to learn class confidence and inter-class relationships more effectively.

Subsequent KD research can be broadly categorized into four families:
(1) Logit-based distillation, which aligns the teacher’s output distribution with the student’s predictions~\cite{chen2020online,jin2023multi,hinton2015distilling,zhao2022decoupled,mirzadeh2020improved,zhang2023not,zhang2018deep,sun2024logit,sun2025knowledge,qian2025good,li2023curriculum,wen2021preparing,cao2023excellent,lan2025improve};
(2) Feature-based distillation, which transfers intermediate representations or attention maps~\cite{ahn2019variational,chen2021distilling,guo2023class,heo2019comprehensive,li2021online,lin2022knowledge,liu2019structured,tian2019contrastive,ji2021show};
(3) Relation-based distillation, which captures structural dependencies between samples or classes~\cite{liu2019knowledge,park2019relational,peng2019correlation}; and
(4) Multi-teacher distillation, where the student aggregates knowledge from multiple teachers~\cite{yuan2021reinforced,zhang2022confidence,son2021densely}.
Feature- and relation-based methods often face challenges due to architectural heterogeneity and the complexity of aligning feature spaces, while multi-teacher approaches increase computational cost and parameterization. Consequently, \emph{logit-based distillation} remains the most practical and widely used paradigm because of its simplicity, generality, and effectiveness.

Within the logit-based family, recent work has focused on refining how the student interacts with the teacher’s softened outputs. Temperature-based methods modify the distillation temperature to control confidence: Li et al.~\cite{li2023curriculum} adapt temperature adversarially in a sample-wise manner, Sun et al.~\cite{sun2024logit} dynamically optimize per-sample temperatures, and Jin et al.~\cite{jin2023multi} exploit multi-level logits for richer supervision. Other approaches attempt to mitigate noisy or overconfident teacher predictions through heuristic corrections~\cite{cao2023excellent,lan2025improve,wen2021preparing,sun2025knowledge}. Despite their empirical gains, these strategies rely heavily on ad-hoc adjustments and hyper-parameter tuning tailored to specific models or datasets. None provide a principled mechanism grounded in statistical theory to handle unreliable teacher outputs.

An orthogonal line of research in \emph{robust statistics} seeks estimators that remain reliable under noisy or contaminated data~\cite{huber2011robust}. Rather than correcting logits heuristically, robustness can be achieved by modifying the underlying loss function so that outliers have a diminished influence. Classical examples include M-estimators~\cite{huber2011robust}, generalized divergence functions~\cite{basu1998robust}, and the \emph{power divergence} family~\cite{cressie1984multinomial,read2012goodness,frydlova2012modified,maji2019robust,chen2004goodness}, which generalizes the KL divergence and introduces a tunable robustness parameter to balance efficiency and noise tolerance. Such divergences have proven effective in diverse settings, including robust variational autoencoders~\cite{akrami2022robust}, outlier-resistant regression~\cite{hernandez2016black}, stabilized GAN training~\cite{cai2020utilizing}, and language model optimization~\cite{roulet2025loss}. However, their integration into deep learning frameworks remains limited and in KD completely unexplored.

Our work bridges knowledge distillation and robust statistics. To the best of our knowledge, this is the first work to provide a unified and statistically grounded framework for \emph{robust knowledge distillation}.

%% file: sec/3_background.tex
\section{Background}
\label{sec:background}

We start by recalling the standard notation and concepts of knowledge
distillation.

Suppose that we have a student model $f_S$ with a parameter vector $\theta$ and our aim is to train it for a $K$-fold classification. In the vanilla setting, we minimize the average  \textit{Kullback--Leibler (KL)} divergence loss between the student prediction and one-hot encoded target
\begin{equation}
\begin{aligned}
\mathcal{L}_{\text{vanilla}}\!\left(\theta, x, y\right)
&\coloneqq\operatorname{KL}\!\left(y, q_\theta\!\left(\cdot|x\right)\right) \\
&= \sum_{k=1}^K y_k \log \frac{y_k}{q_\theta(k\,|\,x)}
\end{aligned}
\label{eq:vanilla_ce}
\end{equation}
over the training dataset consisting of labeled images $\left(x,y\right)\in\mathbb{R}^{3\times H\times W}\otimes\mathbb{R}^K$, where
\begin{align}
q_\theta(k\,|\,x)=\frac{\exp\left({f_S(x, \theta)_k}\right)}{\sum_{d=1}^K\exp\left({f_S(x, \theta)_d}\right)}, \quad k=1,\dots, K
    \label{eq:student_q}
\end{align}
are the student output probabilities. 

In knowledge distillation (\cite{hinton2015distilling}), we work with an enhanced information: a trained teacher model $f_T$ with output probabilities
\begin{align}
    p(k\,|\,x)=\frac{\exp\left({f_T(x)_k}\right)}{\sum_{d=1}^K\exp\left({f_T(x )_d}\right)}, \quad k=1,\dots, K
    \label{eq:teacher_p}
\end{align}
is given and we train the student by also matching its output to the teacher using a loss in the form of a linear combination
\begin{align}
\mathcal{L}_{\text{KD}}\left(\theta,x,y\right)\coloneqq c_1\operatorname{KL}\left(y,q_\theta\left(\cdot|x\right)\right)+c_2\operatorname{KL}\left(p\left(\cdot|x\right),q_\theta\left(\cdot|x\right)\right)
    \label{eq:kd}
\end{align}
for hyper-parameters $c_1,c_2>0$ that balance the influence of ground-truth and teacher predictions. Various modifications to \Cref{eq:kd} have been introduced to further improve the speed and efficiency of knowledge distillation; one of the examples is the Decoupled Knowledge Distillation (\cite{zhao2022decoupled}) where they treat the probabilities corresponding to the ground-truth class separately. As a result in practice, we replace $\operatorname{KL}\!\left(p\!\left(\cdot|x\right), q_\theta\!\left(\cdot|x\right)\right)$ from \Cref{eq:kd} with the following formulation
\begin{equation}
\begin{aligned}
\alpha&\, \operatorname{KL}\!\left(p_{\text{target}}\!\left(\cdot|x\right), q_{\theta,\text{target}}\!\left(\cdot|x\right)\right)
+ \\
&\beta\, \operatorname{KL}\!\left(p_{\text{nontarget}}\!\left(\cdot|x\right), q_{\theta,\text{nontarget}}\!\left(\cdot|x\right)\right),
\end{aligned}
\label{eq:dkd}
\end{equation}
where $\alpha,\beta$ are tunable hyper-parameters that balance the influence of target and non-target classes.

%Additional KD variants have been introduced in the literature and we refer to them in \Cref{sec:experiments}.

%% file: sec/4_methodology.tex
\section{Method}
\label{sec:methodology}

Knowledge distillation typically relies on minimizing the Kullback–Leibler (KL) divergence between teacher and student predictions. Although effective in noise-free regimes (corresponding to a perfect teacher), KL-based training is known to be highly sensitive to distributional misspecification and outliers—issues that arise naturally when the teacher itself is imperfect or overconfident \cite{basu1998robust}.
In this work, we revisit the distillation objective from the perspective of robust statistical estimation. By formulating distillation as a problem of estimating a target distribution under model uncertainty, we connect recent insights from robust divergence estimation in statistics to deep learning objectives. Through this analysis, we identify a principled replacement for the KL divergence: the power divergence, a family that balances efficiency and robustness in a theoretically grounded way.

\subsection{Power-Divergence}

The KL divergence between a data distribution \( p \) and a model distribution \( q_\theta \) can be written as the expected excess ``surprise'' of using \( q_\theta \) to approximate \( p \) where ``surprise'' is measured by the logarithm of the likelihood ratio \( p/q_\theta \):
\begin{equation}
\begin{aligned}
\operatorname{KL}(p, q_\theta)
&= \sum_{k=1}^K p_k \log \frac{p_k}{q_{\theta,k}} \\
&= \mathbb{E}_p \!\left[ \log \!\left( \frac{p}{q_\theta} \right) \right].
\end{aligned}
\label{eq:kl_as_expectation}
\end{equation}
Minimizing this divergence corresponds to maximum likelihood estimation, which yields estimators that are unbiased and efficient~\cite{cramer1999mathematical}.
However, when the data or teacher predictions are contaminated by noise and/or mistakes, such estimators become unstable. 

This motivates the search for \textit{robust alternatives} -- estimators that maintain desirable statistical properties even under model misspecification~\cite{huber2011robust}. We do so by modifying the logarithmic transformation in \Cref{eq:kl_as_expectation}. Ideally, this generalized divergences reduces the influence of large likelihood ratios \(p/q_\theta\) in presence of mistakes in \(p\), thereby limiting sensitivity to outliers~\cite{broniatowski2009parametric,basu1998robust,cressie1984multinomial,pardo2018statistical,read2012goodness}.

\newpage First, we define \(\gamma\)-logarithm (also relaxed logarithm or equivalently, the Box--Cox transform of order \(1-\gamma\)~\cite{cressie1984multinomial,yamano2002some,box1964analysis}) as
\begin{align}
\log_\gamma(x)\coloneqq
\begin{cases}
\log(x), & \text{if } \gamma = 1, \\[6pt]
\dfrac{x^{1 - \gamma} - 1}{1 - \gamma}, & \text{if } \gamma \ne 1, \\[6pt]
\text{undefined}, & \text{if } x \le 0.
\end{cases}
\label{eq:lambda_log}
\end{align}
We have that \(\gamma\)-logarithm continuously approximates the natural logarithm as $\gamma\to 1$ and the tunable parameter $\gamma$ we can control the convexity properties, see \Cref{fig:comparison_plots}.
Substituting the \((1-\lambda)\)-logarithm for the natural logarithm in \Cref{eq:kl_as_expectation} then yields the \emph{power divergence} of order \(\lambda\)
\begin{equation}
\begin{aligned}
\operatorname{D}_{\lambda}(p, q_\theta)
&\coloneqq \frac{1}{1+\lambda} \, \mathbb{E}_p \!\left[\log_{1-\lambda}\!\left(\frac{p}{q_\theta}\right)\right] \\[2mm]
&= \frac{1}{\lambda(\lambda + 1)} \sum_{k=1}^K p(k\,|\,x)
\left[
\left(\frac{p(k\,|\,x)}{q_\theta(k\,|\,x)}\right)^{\lambda}
- 1
\right].
\end{aligned}
\label{eq:power_divergence}
\end{equation}

\input{figures/power_divergence.tex}

This defines a continuous family of \emph{power divergence}~\cite{cressie1984multinomial,amari2012differential} that includes the KL divergence as a special case \(\lambda = 0\), i.e., \(\operatorname{D}_0 = \operatorname{KL}\). 
The parameter \(\lambda\) governs the trade-off between statistical efficiency and robustness: smaller values emphasize accuracy under correct model specification, while larger values attenuate the impact of outliers.
The divergence is convex, nonnegative, and vanishes if and only if \(p = q_\theta\). 
Figure~\ref{fig:comparison_plots} visualizes the behavior of \(\operatorname{D}_\lambda\) as \(\lambda\) varies.

\subsection{Robust Estimator Distillation (REDistill)}

We formulate a robust distillation objective that replaces the KL divergence in the standard formulation with the power divergence of order \(\lambda=2/3\); the choice of \(\lambda\) based on the balance between efficiency and robustness is addressed below.
The resulting loss, which we call \textbf{Robust Estimator Distillation (REDistill)}, is defined as
\begin{equation}
\begin{aligned}
\mathcal{L}_{\text{REDistill}}\left(\theta, x, y\right)
&\coloneqq \operatorname{KL}\!\left(y,\, q_\theta\!\left(\cdot \mid x\right)\right) \\
&\quad + \alpha\, \operatorname{D}_{2/3}\!\left(p_{\text{target}}\!\left(\cdot \mid x\right),\, q_{\theta,\text{target}}\!\left(\cdot \mid x\right)\right) \\
&\quad + \beta\, \operatorname{D}_{2/3}\!\left(p_{\text{nontarget}}\!\left(\cdot \mid x\right),\, q_{\theta,\text{nontarget}}\!\left(\cdot \mid x\right)\right).
\end{aligned}
\label{eq:redistill}
\end{equation}
This formulation inherits the efficiency of KL-based distillation but provides improved robustness to noisy or miscalibrated teacher predictions~\cite{sun2025knowledge}.

To justify our choice of \(\lambda\), we first formally characterize robustness, we analyze the sensitivity of the estimator induced by \(\operatorname{D}_\lambda\) using the \textit{influence function}~\cite{koh2017understanding,huber2011robust} from statistical literature.
Given a loss function \(\mathcal{L}\) and a training set \(\{(x_n, y_n)\}_{n=1}^N\), parameters are estimated as:
\begin{align}
    \hat{\theta}\coloneqq\arg\min_{\theta} \sum_{n=1}^N \mathcal{L}(\theta, x_n, y_n).
\end{align}
When an outlier \((x, y)\) with small weight \(\epsilon>0\) is added, the estimator becomes:
\begin{align}
    \hat{\theta}_{\epsilon, (x, y)}\coloneqq\arg\min_{\theta} \sum_{n=1}^N \mathcal{L}(\theta, x_n, y_n) + \epsilon \mathcal{L}(\theta, x, y).
\end{align}
The influence function quantifies the resulting perturbation:
\begin{align}
    \operatorname{IF}(\mathcal{L}, (x, y))\coloneqq\lim_{\epsilon \to 0^+} \frac{1}{\epsilon} \big(\hat{\theta}_{\epsilon, (x, y)} - \hat{\theta}\big),
    \label{eq:influence_function}
\end{align}
and robustness can be measured by
\begin{align}
    \sup_{(x, y)} \| \operatorname{IF}(\mathcal{L}, (x, y)) \|,
    \label{eq:supremum}
\end{align}
where a smaller value implies greater resistance to outliers.
\input{tables/agnostic}
When training via minimization of $\operatorname{D}_\lambda$, the per-sample loss is
\begin{align}
    \mathcal{L}_\lambda(\theta, x, y) = \operatorname{D}_\lambda\big(y, q_\theta(\cdot | x)\big),
    \label{eq:loss_lambda}
\end{align}
and the corresponding influence function evaluates to (see Supplementary material for details)
\begin{align}
    \operatorname{IF}(\mathcal{L}_\lambda, (x, y)) = \frac{1}{1 + \lambda} \, q_\theta\left(\cdot|x\right).
    \label{eq:if_value}
\end{align}
Equation~\eqref{eq:if_value} demonstrates that larger positive \(\lambda\) values reduce the influence of individual samples, thereby enhancing robustness. 

Following \Cref{eq:if_value} and the analysis of \cite{read2012goodness}, estimators derived from \(\operatorname{D}_\lambda\) remain asymptotically unbiased, with a tunable trade-off: 1) larger $\lambda$ increases robustness to outliers;
    2) smaller $|\lambda|$ improves statistical efficiency. Following theoretical studies~\cite{cressie1984multinomial} we identify $\lambda = 2/3$ as a near-optimal balance between these competing objectives (see Supplementary material), which we adopt in \textit{all our experiments}. %Recent empirical findings in large-scale language modeling~\cite{roulet2025loss} further support this choice.

Unlike previous regularization methods which are mostly heuristic, the robustness in REDistill arises directly from a principled reformulation grounded in robust statistical estimation, without introducing additional hyper-parameters or computational overhead.

\paragraph{Compatibility with Temperature Scaling.} In many practical applications, it is desirable to pre-process model logits by a temperature scaling, that is, we want to soften (possibly) both the teacher \(p\) and model logits \(q_\theta\) by a temperature \(\tau>0\) so that it becomes
\begin{equation}
    \begin{aligned}
    p^\tau(k\,|\,x)&=\frac{\exp\left({f_T(x)_k/\tau}\right)}{\sum_{d=1}^K\exp\left({f_T(x )_d}/\tau\right)}, \\q^\tau(k\,|\,x)&=\frac{\exp\left({f_S(x)_k/\tau}\right)}{\sum_{d=1}^K\exp\left({f_S(x )_d}/\tau\right)}
    \end{aligned}
\end{equation}
However, even at moderate temperatures $\tau$, softening the logits leads to a significant impact on the magnitude of the loss gradient \cite{hinton2015distilling} and therefore slows down the training. We propose that in cases when temperature is used, the power-divergence is scaled with a factor $\tau^2$, i.e. it becomes
\begin{align}
    \tau^2\operatorname{D}_{2/3}\left(p^\tau,q^\tau\right).
\end{align}
See Supplementary material for details.

%% file: figures/power_divergence.tex
\begin{figure}[t]
    \centering

    % Subfigure (a) - power_divergence_plot.pdf
    \begin{subfigure}[b]{0.9\columnwidth}
        \centering
        \includegraphics[width=\columnwidth]{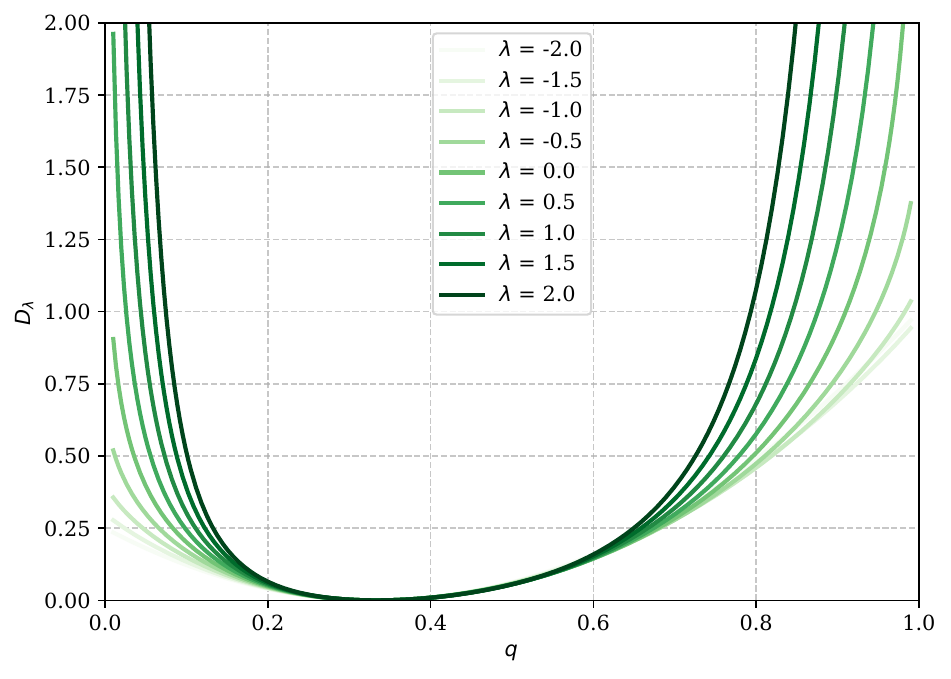}
        %\label{fig:power_divergence}
    \end{subfigure}
    
    \vspace{-15pt}

    % Subfigure (b) - q_log_plot.pdf
    
    \begin{subfigure}[b]{0.9\columnwidth}
        \centering
        \includegraphics[width=\columnwidth]{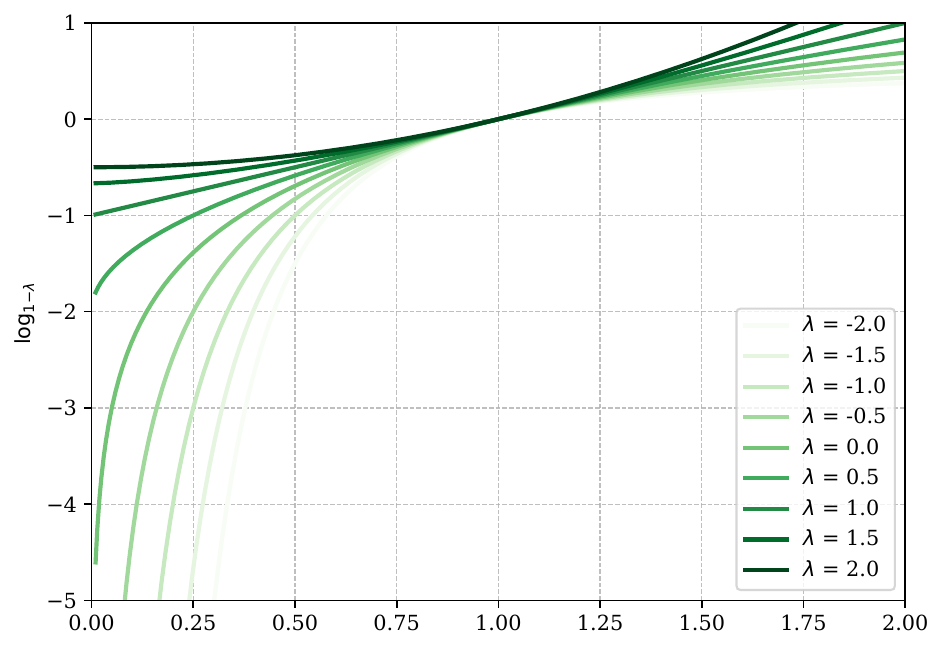}
        %\label{fig:q_log}
        \vspace{-15pt}
    \end{subfigure}        
    \caption{The divergence $\operatorname{D}_\lambda$ corresponds to the $\operatorname{KL}$ divergence for $\lambda=0$. For other values, logarithm as a measure of surprise in the divergence computation is replaced by its smooth relaxation (known as $\left(1-\lambda\right)$~-logarithm, see \cref{eq:lambda_log}). (a) shows graph of $\left(1-\lambda\right)$~-logarithm, (b) shows $\operatorname{D}_\lambda(P \| Q)$ between $P=\left(1/3,2/3\right)$ and $Q=\left(q, 1-q\right)$ as a function of $q$ for different $\lambda$ values.}
    \label{fig:comparison_plots}
\end{figure}

%% file: tables/agnostic.tex
\begin{table*}
    \centering    
    \small
\begin{tabular}{l|ccccccc}

teacher & \begin{tabular}{@{}c@{}}ResNet32×4 \\ 79.42\end{tabular} & \begin{tabular}{@{}c@{}}ResNet32×4 \\ 79.42\end{tabular} & \begin{tabular}{@{}c@{}}ResNet32×4 \\ 79.42\end{tabular} & \begin{tabular}{@{}c@{}}WRN-40-2 \\ 75.61\end{tabular} & \begin{tabular}{@{}c@{}}WRN-40-2 \\ 75.61\end{tabular} & \begin{tabular}{@{}c@{}}VGG13 \\ 74.64\end{tabular} & \begin{tabular}{@{}c@{}}ResNet50 \\ 79.34\end{tabular} \\
student & \begin{tabular}{@{}c@{}}SHN-V2 \\ 71.82\end{tabular} & \begin{tabular}{@{}c@{}}WRN-16-2 \\ 73.26\end{tabular} & \begin{tabular}{@{}c@{}}WRN-40-2 \\ 75.61\end{tabular} & \begin{tabular}{@{}c@{}}ResNet8×4 \\ 72.50\end{tabular} & \begin{tabular}{@{}c@{}}MN-V2 \\ 64.60\end{tabular} & \begin{tabular}{@{}c@{}}MN-V2 \\ 64.60\end{tabular} & \begin{tabular}{@{}c@{}}MN-V2 \\ 64.60\end{tabular} \\
\hline
KD~\cite{hinton2015distilling} & 75.51 & 72.93 & 77.15 & 75.03 & 64.66 & 64.23 & 64.40 \\
DKD~\cite{zhao2022decoupled} & 77.01 & 76.18 & 78.81 & 75.50 & 69.27 & 69.70 & 70.60 \\
LSKD~\cite{sun2024logit} & 77.11 & 76.25 & 78.86 & 76.55 & 69.57 & 69.82 & 70.61 \\
RLD~\cite{sun2025knowledge} & 76.74 & 75.97 & 78.68 & 75.12 & 68.91 & 69.72 & 70.22 \\
\textit{REDistill (ours)} & \textbf{77.52} & \textbf{76.31} & \textbf{79.01} & \textbf{77.21} & \textbf{69.98} & \textbf{70.12} & \textbf{70.92} \\
\hline \hline
teacher & \begin{tabular}{@{}c@{}}ResNet32×4 \\ 79.42\end{tabular} & \begin{tabular}{@{}c@{}}VGG13 \\ 74.64\end{tabular} & \begin{tabular}{@{}c@{}}WRN-40-2 \\ 75.61\end{tabular} & \begin{tabular}{@{}c@{}}WRN-40-2 \\ 75.61\end{tabular} & \begin{tabular}{@{}c@{}}ResNet56 \\ 72.34\end{tabular} & \begin{tabular}{@{}c@{}}ResNet110 \\ 74.31\end{tabular} & \begin{tabular}{@{}c@{}}ResNet110 \\ 74.31\end{tabular} \\
student & \begin{tabular}{@{}c@{}}ResNet8×4 \\ 72.50\end{tabular} & \begin{tabular}{@{}c@{}}VGG8 \\ 70.36\end{tabular} & \begin{tabular}{@{}c@{}}WRN-40-1 \\ 71.98\end{tabular} & \begin{tabular}{@{}c@{}}WRN-16-2 \\ 73.26\end{tabular} & \begin{tabular}{@{}c@{}}ResNet20 \\ 69.06\end{tabular} & \begin{tabular}{@{}c@{}}ResNet32 \\ 71.14\end{tabular} & \begin{tabular}{@{}c@{}}ResNet20 \\ 69.06\end{tabular} \\
\hline
KD~\cite{hinton2015distilling} & 73.94 & 70.69 & 71.40 & 73.83 & 71.01 & 72.17 & 69.95 \\
DKD~\cite{zhao2022decoupled} & 76.17 & 74.58 & 74.57 & 75.56 & 71.71 & 73.53 & 71.64 \\
LSKD~\cite{sun2024logit} & 76.67 & 74.61 & 74.88 & 75.55 & 71.51 & 73.94 & \textbf{71.67} \\
RLD~\cite{sun2025knowledge} & 76.12 & 74.40 & 74.69 & 75.62 & 70.85 & 73.09 & 70.18 \\
%AID\cite{qian2025good} & 1 & 1 & 1 & 1 & 1 & 1 & 1 \\
\textit{REDistill (ours)} & \textbf{77.05} & \textbf{74.78} & \textbf{74.92} & \textbf{76.82} & \textbf{71.95} & \textbf{74.18} & 71.60 \\

\end{tabular}
\caption{Student network top-1 accuracy (\%) on the CIFAR-100 validation set, using the \textit{model-agnostic} hyper-parameters protocol (see \cref{subsec:setup}). The reported results are averages of four trials.\vspace{-10pt}}
\label{tab:table_agnostic}
\end{table*}

% Runs from Nov 3rd 2025 for KD, DKD, LSKD from cifar100_baselines project, for RLD we use runs named *reproduce* from renyi-kd-new from 20th October 2025

%% file: sec/5_experiments.tex
\section{Experiments}
\label{sec:experiments}
\input{tables/specific}

\subsection{Experimental Setup}
\label{subsec:setup}
In line with previous work, we conduct experiments on two standard image classification benchmarks: CIFAR-100 \cite{krizhevsky2009learning} and ImageNet-1k \cite{russakovsky2015imagenet}, with input resolutions of \(32\times32\) and \(224\times224\), respectively. The teacher–student model pairs include variants of ResNet \cite{he2016deep}, WideResNet (WRN) \cite{zagoruyko2016wide}, VGG \cite{simonyan2014very}, ShuffleNet (SHN-V2) \cite{ma2018shufflenet,zhang2018shufflenet} and MobileNetV2 (MN-V2) \cite{howard2017mobilenets,sandler2018mobilenetv2}.

All the compared KD losses for different methods depend on various hyper-parameter settings (i.e. logit temperature and $c_1,c_2$ in \Cref{eq:kd} for KD \cite{hinton2015distilling}, $\alpha,\beta$ in \Cref{eq:dkd} in DKD \cite{zhao2022decoupled}). First, we compare all methods in \textbf{model-agnostic evaluation protocol}, where all \textit{hyper-parameter values are fixed for a given method} for all evaluated models. We argue that this setting is more realistic, because it is impractical to assume that one has to find optimal hyper-parameter setting for each teacher-student pair on every dataset. It also makes the comparison between methods more fair, because there is no possibility of hyper-parameter tuning on the validation set to artificially boost accuracy of given method. In contrast, a model-specific protocol is useful when the goal is to determine the best achievable performance for each method.

\input{tables/imagenet}
\input{tables/combined}

Second, in line with previous work~\cite{sun2024logit,zhao2022decoupled,sun2025knowledge}, we also include \textbf{model-specific evaluation protocol}, where each teacher-student pair has its own unique hyper-parameter values. While model-specific hyper-parameters yield slightly better results and are important for assessment of the best possible student accuracy, training for a single teacher-student pair means training for each possible hyper-parameter combination separately, which leads to extensive computation demands as the optimal hyper-parameters are unknown and need to be searched for. 
% \begin{enumerate}
%     \item \textbf{Model-specific hyper-parameters} — following previous work , the temperature and distillation loss weights are individually tuned for each teacher–student pair.
%     \item \textbf{Model-agnostic hyper-parameters} — to better assess generalization across diverse architectures, all hyper-parameters are fixed and shared across teacher-student pairs.
% \end{enumerate}
% In contrast with previous works, we include both the protocols.  This is why we included also Protocol~2 which is more realistic when the method is applied for a new teacher-student pair.

Our comparisons focus on logit-based knowledge distillation approaches KD \cite{hinton2015distilling}, CTKD \cite{li2023curriculum}, DKD \cite{zhao2022decoupled}, LSKD \cite{sun2024logit}, LA \cite{wen2021preparing}, RC \cite{cao2023excellent}, LR \cite{lan2025improve}, and RLD \cite{sun2025knowledge}. For feature-based methods, we include FitNet \cite{romero2015fitnets}, AT \cite{zagoruyko2016paying}, RKD \cite{park2019relational}, CRD \cite{tian2019contrastive}, OFD \cite{heo2019comprehensive}, ReviewKD \cite{chen2021distilling}, SimKD \cite{chen2022knowledge}, CAT-KD \cite{guo2023class}. These methods are compared separately from logit-based methods as they are more computationally demanding \cite{chen2022knowledge,sun2025knowledge} and less generalizable across heterogeneous architectures \cite{wang2022semckd}.%, and often yield comparable or inferior performance \cite{sun2024logit,sun2025knowledge}.

All results were obtained following an extensive review of prior work and publicly available source codes to ensure fair and consistent comparisons. 
%We exclude MLKD \cite{jin2023multi} due to incompatible experimental settings
Furthermore, we note that AID and SKD \cite{qian2025good,yuan2024student} are not included (same as  \cite{sun2025knowledge}), as both involve teacher fine-tuning either before distillation (AID) or during distillation through an attention head (SKD), making their teacher performance incomparable with other methods. For \textit{model-agnostic evaluation}, we reproduce all results using the authors’ official code, using the default/best reported hyper-parameters from official implementations. For our method, we used \(\lambda=2/3\) and used the $\alpha$ and $\beta$ values from DKD~\cite{sun2024logit}. For \textit{model-specific evaluation}, the results are taken directly as reported in \cite{sun2024logit,sun2025knowledge}. 

In both cases, we follow the same training setup as in \cite{sun2024logit,zhao2022decoupled,sun2025knowledge}. For CIFAR-100, we train with SGD \cite{sutskever2013importance} for 240 epochs with a batch size of 64, weight decay of $5\times10^{-4}$, and momentum 0.9. The initial learning rate is 0.01 for SHN-V2 and MN-V2 and 0.05 for the other architectures; it is reduced by a factor of 10 at epochs 150, 180, and 210.
For ImageNet, we train with SGD \cite{sutskever2013importance} using a batch size of 512, 100 epochs, weight decay $1\times10^{-4}$, momentum 0.9, and initial learning rate 0.2, decayed by 10 every 30 epochs.  
Unlike \cite{sun2025knowledge}, we do not use any data augmentation. All experiments were conducted on single Tesla V100-SXM2-32GB GPU; one training on CIFAR-100 required up to 2 hours, while ImageNet training took \(\sim 2\) days.

\input{tables/feature}

\subsection{Results}
For CIFAR-100 and ImageNet-1k, we present results for both \textit{model-agnostic} (\Cref{tab:table_agnostic}) and \textit{model-specific} (\Cref{tab:table_specific,tab:imagenet_results}) protocols. \textit{Our method consistently surpasses all previous knowledge distillation methods}, with the exception of the VGG13/VGG8 pair where it places second. On ImageNet-1k (\Cref{tab:imagenet_results}), our method achieves \textbf{state-of-the-art accuracy} with a substantial margin over the best existing baselines.

These results confirm our central hypothesis: incorporating a robust training loss into knowledge distillation leads to stronger student models, and the gains are not attributable to hyper-parameter tuning. Moreover, comparing results from the \textit{model-agnostic} and \textit{model-specific} settings reveals that several high-performing methods in \Cref{tab:table_specific} rely on extensive hyper-parameter tuning and/or data augmentation -- without it, their performance drops significantly. For example, RLD underperforms even the standard KD baseline for ResNet56/ResNet20 and yields only marginal gains ($\sim0.2\%$) for ResNet110/ResNet20 and WRN-40-2/ResNet8x4 -- an order of magnitude smaller than under the \textit{model-specific} protocol. Similarly, LSKD does not consistently outperform DKD. In contrast, our method maintains strong and stable improvements in both evaluation settings, demonstrating its robustness and generalization across diverse teacher–student architectures.

\paragraph{Combining methods.} Next, we demonstrate the extensive applicability of our method by incorporating it into existing distillation pipelines within \textit{model-specific} protocol to demonstrate the results improvements.  We integrate it with three representative approaches -- KD~\cite{hinton2015distilling}, DKD~\cite{zhao2022decoupled}, and MLKD~\cite{jin2023multi} -- without altering their training pipelines. As seen in \Cref{tab:table_combined}, adding our loss leads to further improvement of the student model in every case, indicating that the robustness introduced by our loss complements rather than competes with prior formulations.
Notably, combining our method with MLKD yields performance that exceeds feature-based distillation methods \Cref{tab:table_feature}, despite being significantly simpler and requiring no additional feature matching or architectural modifications. This highlights that our contribution is orthogonal to existing strategies: it enhances them by providing a more stable and reliable training signal.
%Overall, our method is (i) effective, outperforming logit-based baselines; (ii) robust, maintaining gains without hyper-parameter tuning or augmentation; and (iii) easy to plug-in, providing consistent improvements when combined with other distillation objectives.

\subsection{Ablations}
\paragraph{Robustness parameter.} In the ablation study, we investigate the impact of $\lambda$ on the performance of our method (see \Cref{eq:power_divergence}). Recall that $\lambda$ is not treated as a hyper-parameter, but is fixed at $2/3$ based on theoretical evidence. As shown in \Cref{tab:ablation_lambda}, we evaluate top-1 accuracy on the CIFAR-100 validation set for various values of $\lambda$, ranging from $0$ to $2$. The value of $\lambda$ controls the relative weighting between the teacher's and student's predictions during the distillation process.
The results show that values around $\lambda = 2/3$ yield the highest top-1 accuracy. This supports our theoretical argument that $\lambda = 2/3$ provides an optimal balance between the teacher's and student's knowledge, resulting in a more robust distillation process. Furthermore, the choice of $\lambda$ appears somewhat flexible, with other values in this region also yielding strong performance. However, when $\lambda$ exceeds $1$, the performance starts to degrade. Specifically, $\lambda = 3/2$ and $\lambda = 2$ lead to lower accuracies.

\input{tables/ablation_lambda}

%% file: tables/specific.tex
\begin{table*}
    \centering    
    \small
\begin{tabular}{l|ccccccc}

teacher & \begin{tabular}{@{}c@{}}ResNet32×4 \\ 79.42\end{tabular} & \begin{tabular}{@{}c@{}}ResNet32×4 \\ 79.42\end{tabular} & \begin{tabular}{@{}c@{}}ResNet32×4 \\ 79.42\end{tabular} & \begin{tabular}{@{}c@{}}WRN-40-2 \\ 75.61\end{tabular} & \begin{tabular}{@{}c@{}}WRN-40-2 \\ 75.61\end{tabular} & \begin{tabular}{@{}c@{}}VGG13 \\ 74.64\end{tabular} & \begin{tabular}{@{}c@{}}ResNet50 \\ 79.34\end{tabular} \\
student & \begin{tabular}{@{}c@{}}SHN-V2 \\ 71.82\end{tabular} & \begin{tabular}{@{}c@{}}WRN-16-2 \\ 73.26\end{tabular} & \begin{tabular}{@{}c@{}}WRN-40-2 \\ 75.61\end{tabular} & \begin{tabular}{@{}c@{}}ResNet8×4 \\ 72.50\end{tabular} & \begin{tabular}{@{}c@{}}MN-V2 \\ 64.60\end{tabular} & \begin{tabular}{@{}c@{}}MN-V2 \\ 64.60\end{tabular} & \begin{tabular}{@{}c@{}}MN-V2 \\ 64.60\end{tabular} \\
\hline
KD~\cite{hinton2015distilling} & 74.45 & 74.90 & 77.70 & 73.97 & 68.36 & 67.37 & 67.35 \\
CTKD~\cite{li2023curriculum} & 75.37 & 74.57 & 77.66 & 74.61 & 68.34 & 68.50 & 68.67 \\
DKD~\cite{zhao2022decoupled} & 77.07 & 75.70 & 78.46 & 75.56 & 69.28 & 69.71 & 70.35 \\
LSKD~\cite{sun2024logit} & 77.37 & 76.19 & \textit{78.95} & \textit{76.75} & \textit{70.01} & \textit{69.98} & 70.45 \\
LA\cite{wen2021preparing} & 75.14 & -- & 77.39 & 73.88 & 68.57 & 68.09 & 68.85 \\
RC~\cite{cao2023excellent} & 75.61 & -- & 77.58 & 75.22 & 68.72 & 68.66 & 68.98 \\
LR~\cite{lan2025improve} & 76.27 & -- & 78.73 & 75.26 & 69.02 & 69.78 & 70.38 \\
%SKD\cite{yuan2024student} & 76.96 & -- & -- & -- & -- & -- & 69.99 \\
RLD~\cite{sun2025knowledge}\dag & \textit{77.56} & -- & 78.91 & 76.12 & 69.75 & 69.97 & \textit{70.76} \\
%AID\cite{qian2025good} & 1 & 1 & 1 & 1 & 1 & 1 & 1 \\
\textit{REDistill (ours)} & \textbf{77.84} & \textbf{76.37} & \textbf{79.23} & \textbf{77.46} & \textbf{70.48} & \textbf{70.58} & \textbf{71.03} \\
\hline \hline
teacher & \begin{tabular}{@{}c@{}}ResNet32×4 \\ 79.42\end{tabular} & \begin{tabular}{@{}c@{}}VGG13 \\ 74.64\end{tabular} & \begin{tabular}{@{}c@{}}WRN-40-2 \\ 75.61\end{tabular} & \begin{tabular}{@{}c@{}}WRN-40-2 \\ 75.61\end{tabular} & \begin{tabular}{@{}c@{}}ResNet56 \\ 72.34\end{tabular} & \begin{tabular}{@{}c@{}}ResNet110 \\ 74.31\end{tabular} & \begin{tabular}{@{}c@{}}ResNet110 \\ 74.31\end{tabular} \\
student & \begin{tabular}{@{}c@{}}ResNet8×4 \\ 72.50\end{tabular} & \begin{tabular}{@{}c@{}}VGG8 \\ 70.36\end{tabular} & \begin{tabular}{@{}c@{}}WRN-40-1 \\ 71.98\end{tabular} & \begin{tabular}{@{}c@{}}WRN-16-2 \\ 73.26\end{tabular} & \begin{tabular}{@{}c@{}}ResNet20 \\ 69.06\end{tabular} & \begin{tabular}{@{}c@{}}ResNet32 \\ 71.14\end{tabular} & \begin{tabular}{@{}c@{}}ResNet20 \\ 69.06\end{tabular} \\
\hline
KD~\cite{hinton2015distilling} & 73.33 & 72.98 & 73.54 & 74.92 & 70.66 & 73.08 & 70.67 \\
CTKD~\cite{li2023curriculum} & 73.39 & 73.52 & 73.93 & 75.45 & 71.19 & 73.52 & 70.99 \\
DKD~\cite{zhao2022decoupled} & 76.32 & 74.68 & 74.81 & 76.24 & 71.97 & 74.11 & 71.06 \\
LSKD~\cite{sun2024logit} & 77.01 & 74.81 & 74.89 & 76.39 & 72.32 & 74.29 & 71.85 \\
LA~\cite{wen2021preparing} & 73.46 & 73.51 & 73.75 & -- & 71.24 & 73.39 & 70.86 \\
RC~\cite{cao2023excellent} & 74.68 & 73.37 & 74.07 & -- & 71.63 & 73.44 & 71.41 \\
LR~\cite{lan2025improve} & 76.06 & 74.66 & 74.42 & -- & 70.74 & 73.52 & 70.61 \\
%SKD\cite{yuan2024student} & 76.84 & \textbf{74.94} & 74.52 & 76.29 & 71.73 & 74.06 & -- \\
RLD~\cite{sun2025knowledge} & 76.64 & \textbf{74.93} & 74.88 & -- & 72.00 & 74.02 & 71.67 \\
%AID\cite{qian2025good} & 1 & 1 & 1 & 1 & 1 & 1 & 1 \\
\textit{REDistill (ours)} & \textbf{77.31} & 74.88 & \textbf{74.97} & \textbf{76.99} & \textbf{72.41} & \textbf{74.40} & \textbf{71.89} \\
\end{tabular}
\caption{Student network top-1 accuracy (\%) on the CIFAR-100 validation set, using the \textit{model-specific} hyper-parameters protocol (see \cref{subsec:setup}). The reported results are averages of four trials.}
\label{tab:table_specific}
\end{table*}

%% file: tables/imagenet.tex
\begin{table}
\centering
\small
\begin{tabular}{lcccc}
& \multicolumn{2}{c}{Res34/Res18} & \multicolumn{2}{c}{Res50/MN-V1} \\
\cline{2-5}
 & Top-1 & Top-5 & Top-1 & Top-5 \\
\hline
Teacher & 73.31 & 91.42 & 76.16 & 92.86 \\
Student & 69.75 & 89.07 & 68.87 & 88.76 \\
\hline
KD~\cite{hinton2015distilling} & 71.03 & 90.05 & 70.50 & 89.80 \\
CTKD~\cite{li2023curriculum} & 71.38 & 90.27 & 71.16 & 90.11 \\
DKD~\cite{zhao2022decoupled} & 71.70 & 90.41 & 72.05 & 91.05 \\
LSKD~\cite{sun2024logit} & 71.88 & 90.58 & 72.85 & 91.23 \\
LA~\cite{wen2021preparing} & 71.17 & 90.16 & 70.98 & 90.13 \\
RC~\cite{cao2023excellent} & 71.59 & 90.21 & 71.86 & 90.54 \\
LR~\cite{sun2024logit} & 70.29 & 89.98 & 71.76 & 90.93 \\
RLD~\cite{sun2025knowledge} & 71.91 & 90.59 & 72.75 & 91.18 \\
\hline
\textit{REDistill (ours)} & \textbf{72.00} & \textbf{90.71} & \textbf{72.98} & \textbf{91.48} \\
\end{tabular}
\caption{Student network accuracy (\%) on the ImageNet-1k validation set. The reported results are averages of three trials.}
\label{tab:imagenet_results}
\end{table}

%% file: tables/combined.tex
\begin{table*}[t]
    \centering    
    \small

\begin{tabular}{l|ccccccc}

teacher & \begin{tabular}{@{}c@{}}ResNet32×4 \\ 79.42\end{tabular} & \begin{tabular}{@{}c@{}}ResNet32×4 \\ 79.42\end{tabular} & \begin{tabular}{@{}c@{}}ResNet32×4 \\ 79.42\end{tabular} & \begin{tabular}{@{}c@{}}WRN-40-2 \\ 75.61\end{tabular} & \begin{tabular}{@{}c@{}}WRN-40-2 \\ 75.61\end{tabular} & \begin{tabular}{@{}c@{}}VGG13 \\ 74.64\end{tabular} & \begin{tabular}{@{}c@{}}ResNet50 \\ 79.34\end{tabular} \\
student & \begin{tabular}{@{}c@{}}SHN-V2 \\ 71.82\end{tabular} & \begin{tabular}{@{}c@{}}WRN-16-2 \\ 73.26\end{tabular} & \begin{tabular}{@{}c@{}}WRN-40-2 \\ 75.61\end{tabular} & \begin{tabular}{@{}c@{}}ResNet8×4 \\ 72.50\end{tabular} & \begin{tabular}{@{}c@{}}MN-V2 \\ 64.60\end{tabular} & \begin{tabular}{@{}c@{}}MN-V2 \\ 64.60\end{tabular} & \begin{tabular}{@{}c@{}}MN-V2 \\ 64.60\end{tabular} \\
\hline
KD~\cite{hinton2015distilling} & 74.45 & 74.90 & 77.70 & 73.97 & 68.36 & 67.37 & 67.35 \\
REDistill+KD (ours) & 74.99 & 75.62 & 77.85 & 74.20 & 68.69 & 67.52 & 68.02 \\
DKD~\cite{zhao2022decoupled} & 77.07 & 75.70 & 78.46 & 75.56 & 69.28 & 69.71 & 70.35 \\
REDistill+DKD (ours) & 77.84 & 76.37 & 79.23 & 77.46 & 70.48 & 70.58 & 71.03 \\
%RLD\cite{sun2025knowledge} & 77.56 & -- & 78.91 & 76.12 & 69.75 & 69.97 & 70.76 \\
%RED+RLD (ours) & X & X & X & X & X & X & X \\
MLKD~\cite{jin2023multi} & 78.44 & 76.52 & 79.26 & 77.33 & 70.78 & 70.57 & 71.04\\ 
\textit{REDistill+MLKD (ours)} & \textbf{78.82} & \textbf{77.50} & \textbf{79.60} & \textbf{77.63} & \textbf{71.63} & \textbf{70.95} & \textbf{71.20} \\
\hline \hline
teacher & \begin{tabular}{@{}c@{}}ResNet32×4 \\ 79.42\end{tabular} & \begin{tabular}{@{}c@{}}VGG13 \\ 74.64\end{tabular} & \begin{tabular}{@{}c@{}}WRN-40-2 \\ 75.61\end{tabular} & \begin{tabular}{@{}c@{}}WRN-40-2 \\ 75.61\end{tabular} & \begin{tabular}{@{}c@{}}ResNet56 \\ 72.34\end{tabular} & \begin{tabular}{@{}c@{}}ResNet110 \\ 74.31\end{tabular} & \begin{tabular}{@{}c@{}}ResNet110 \\ 74.31\end{tabular} \\
student & \begin{tabular}{@{}c@{}}ResNet8×4 \\ 72.50\end{tabular} & \begin{tabular}{@{}c@{}}VGG8 \\ 70.36\end{tabular} & \begin{tabular}{@{}c@{}}WRN-40-1 \\ 71.98\end{tabular} & \begin{tabular}{@{}c@{}}WRN-16-2 \\ 73.26\end{tabular} & \begin{tabular}{@{}c@{}}ResNet20 \\ 69.06\end{tabular} & \begin{tabular}{@{}c@{}}ResNet32 \\ 71.14\end{tabular} & \begin{tabular}{@{}c@{}}ResNet20 \\ 69.06\end{tabular} \\
\hline 
KD~\cite{hinton2015distilling} & 73.33 & 72.98 & 73.54 & 74.92 & 70.66 & 73.08 & 70.67 \\
REDistill+KD (ours) & 74.12 & 73.41 & 74.62 & 75.25 & 71.01 & 73.25 & 70.89 \\
DKD~\cite{zhao2022decoupled} & 76.32 & 74.68 & 74.81 & 76.24 & 71.97 & 74.11 & 71.06 \\
REDistill+DKD (ours) & 77.31 & 74.88 & 74.97 & 76.99 & 72.41 & 74.40 & 71.89 \\
%RLD\cite{sun2025knowledge} & 76.64 & 74.93 & 74.88 & -- & 72.00 & 74.02 & 71.67 \\
%RED+RLD (ours) & X & X & X & X & X & X & X \\
MLKD~\cite{jin2023multi} & 77.08 & 75.18 & 75.35 & 76.63 & 72.19 & 74.11 & 71.89\\
\textit{REDistill+MLKD (ours)} & \textbf{78.31} & \textbf{75.24} & \textbf{75.53} & \textbf{76.91} & \textbf{72.40} & \textbf{74.29} & \textbf{72.23} \\
\end{tabular}
\caption{Combining REDistill with existing methods. Student network top-1 accuracy (\%) on the CIFAR-100 validation set.\vspace{-10pt}}
\label{tab:table_combined}
\end{table*}

%% file: tables/feature.tex
\begin{table*}[t]
    \centering    
    \small
\begin{tabular}{l|ccccccc}

teacher & \begin{tabular}{@{}c@{}}ResNet32×4 \\ 79.42\end{tabular} & \begin{tabular}{@{}c@{}}ResNet32×4 \\ 79.42\end{tabular} & \begin{tabular}{@{}c@{}}ResNet32×4 \\ 79.42\end{tabular} & \begin{tabular}{@{}c@{}}WRN-40-2 \\ 75.61\end{tabular} & \begin{tabular}{@{}c@{}}WRN-40-2 \\ 75.61\end{tabular} & \begin{tabular}{@{}c@{}}VGG13 \\ 74.64\end{tabular} & \begin{tabular}{@{}c@{}}ResNet50 \\ 79.34\end{tabular} \\
student & \begin{tabular}{@{}c@{}}SHN-V2 \\ 71.82\end{tabular} & \begin{tabular}{@{}c@{}}WRN-16-2 \\ 73.26\end{tabular} & \begin{tabular}{@{}c@{}}WRN-40-2 \\ 75.61\end{tabular} & \begin{tabular}{@{}c@{}}ResNet8×4 \\ 72.50\end{tabular} & \begin{tabular}{@{}c@{}}MN-V2 \\ 64.60\end{tabular} & \begin{tabular}{@{}c@{}}MN-V2 \\ 64.60\end{tabular} & \begin{tabular}{@{}c@{}}MN-V2 \\ 64.60\end{tabular} \\
\hline
FitNet \cite{romero2015fitnets} & 73.54 & 74.70 & 77.69 & 74.61 & 68.64 & 64.16 & 63.16 \\
AT \cite{zagoruyko2016paying} & 72.73 & 73.91 & 77.43 & 74.11 & 60.78 & 59.40 & 58.58 \\
RKD \cite{park2019relational} & 73.21 & 74.86 & 77.82 & 75.26 & 69.27 & 64.52 & 64.43 \\
CRD \cite{tian2019contrastive} & 75.65 & 75.65 & 78.15 & 75.24 & 70.28 & 69.73 & 69.11 \\
OFD \cite{heo2019comprehensive} & 76.82 & 76.17 & 79.25 & 74.36 & 69.92 & 69.48 & 69.04 \\
ReviewKD \cite{chen2021distilling} & 77.78 & 76.11 & 78.96 & 74.34 & 71.28 & 70.37 & 69.89 \\
SimKD \cite{chen2022knowledge} & 78.39 & 77.17 & 79.29 & 75.29 & 70.10 & 69.44 & 69.97 \\
CAT-KD \cite{guo2023class} & 78.41 & 76.97 & 78.59 & 75.38 & 70.24 & 69.13 & \textbf{71.36} \\  \hline
\textit{REDistill+MLKD (ours)} & \textbf{78.82} & \textbf{77.50} & \textbf{79.60} & \textbf{77.63} & \textbf{71.63} & \textbf{70.95} & 71.20 \\
\hline \hline
teacher & \begin{tabular}{@{}c@{}}ResNet32×4 \\ 79.42\end{tabular} & \begin{tabular}{@{}c@{}}VGG13 \\ 74.64\end{tabular} & \begin{tabular}{@{}c@{}}WRN-40-2 \\ 75.61\end{tabular} & \begin{tabular}{@{}c@{}}WRN-40-2 \\ 75.61\end{tabular} & \begin{tabular}{@{}c@{}}ResNet56 \\ 72.34\end{tabular} & \begin{tabular}{@{}c@{}}ResNet110 \\ 74.31\end{tabular} & \begin{tabular}{@{}c@{}}ResNet110 \\ 74.31\end{tabular} \\
student & \begin{tabular}{@{}c@{}}ResNet8×4 \\ 72.50\end{tabular} & \begin{tabular}{@{}c@{}}VGG8 \\ 70.36\end{tabular} & \begin{tabular}{@{}c@{}}WRN-40-1 \\ 71.98\end{tabular} & \begin{tabular}{@{}c@{}}WRN-16-2 \\ 73.26\end{tabular} & \begin{tabular}{@{}c@{}}ResNet20 \\ 69.06\end{tabular} & \begin{tabular}{@{}c@{}}ResNet32 \\ 71.14\end{tabular} & \begin{tabular}{@{}c@{}}ResNet20 \\ 69.06\end{tabular} \\
\hline
FitNet \cite{romero2015fitnets} & 73.50 & 71.02 & 72.24 & 73.58 & 69.21 & 71.06 & 68.99 \\
AT \cite{zagoruyko2016paying} & 73.44 & 71.43 & 72.77 & 74.08 & 70.55 & 72.31 & 70.65 \\
RKD \cite{park2019relational} & 71.90 & 71.48 & 72.22 & 73.35 & 69.61 & 71.82 & 69.25 \\
CRD \cite{tian2019contrastive} & 75.51 & 73.94 & 74.14 & 75.48 & 71.16 & 73.48 & 71.46 \\
OFD \cite{heo2019comprehensive} & 74.95 & 73.95 & 74.33 & 75.24 & 70.98 & 73.23 & 71.29 \\
ReviewKD \cite{chen2021distilling} & 75.63 & 74.84 & 75.09 & 76.12 & 71.89 & 73.89 & 71.34 \\
SimKD \cite{chen2022knowledge} & 78.08 & 74.89 & 74.53 & 75.53 & 71.05 & 73.92 & 71.06 \\
CAT-KD \cite{guo2023class} & 76.91 & 74.65 & 74.82 & 75.60 & 71.62 & 73.62 & 71.37 \\ \hline
\textit{REDistill+MLKD (ours)} & \textbf{78.31} & \textbf{75.24} & \textbf{75.53} & \textbf{76.91} & \textbf{72.40} & \textbf{74.29} & \textbf{72.23} \\
\end{tabular}
\caption{Comparison with feature-based KD methods. Student network top-1 accuracy (\%) on the CIFAR-100 validation set. \vspace{-10pt} }
\label{tab:table_feature}
\end{table*}

%% file: tables/ablation_lambda.tex
\begin{table}
\centering
\small
\begin{tabular}{lc}
\hline
\multicolumn{1}{c}{$\lambda$} & \multicolumn{1}{c}{Top-1 Accuracy (\%)} \\
\hline
$0$         & 77.11 \\
$1/3$ & 77.12 \\
$1/2$ & 77.32 \\
$2/3$ (ours) & \textbf{77.52} \\
$1$           & 77.50 \\
$3/2$ & 77.01 \\
$2$           & 76.31 \\
\hline
\end{tabular}
\caption{Student network accuracy on the CIFAR-100 validation dataset for different $\lambda$ values with teacher/student pair ResNet32x4/SHN-V2. Note that $\lambda=0$ corresponds to DKD~\cite{zhao2022decoupled}.\vspace{-10pt}}
\label{tab:ablation_lambda}
\end{table}

%% file: sec/6_conclusion.tex
\section{Conclusion}
\label{sec:conclusion}

In this work, we revisited the robustness of knowledge distillation (KD) through the lens of robust statistical estimation. We identify a key weakness in conventional KD -- the assumption that teacher soft targets are always reliable. When teachers are imperfect or overconfident, the standard KL divergence becomes overly sensitive to noise, leading to unstable or suboptimal student training.
To address this, we propose Robust Estimator Distillation (REDistill), which replaces the KL divergence with a power divergence. This yields a principled loss that reduces unreliable teacher output while preserving informative structure -- \textit{without heuristics, extra hyper-parameters, architectural changes, or additional computational cost}.
Experiments on CIFAR-100 and ImageNet showed that REDistill consistently improved student performance and achieves \textit{state-of-the-art results} while (i) being effective and outperforming logit-based baselines; (ii) not being sensitive to hyper-parameter tuning and/or data augmentation -- an important property for generalization to new teacher-student pairs; and (iii) \textit{easy to plug in}, providing consistent improvements when combined with other distillation objectives, showing that our contribution is orthogonal to existing strategies.
%In general, REDistill grounds KD in robust estimation theory, offering a simple, interpretable, and effective framework for training reliable students under imperfect teachers by providing a more stable and reliable training signal.

%% file: sec/X_suppl.tex
\clearpage
%\setcounter{page}{1}
%\maketitlesupplementary
\twocolumn[
\centering
\Large
\vspace{0.5em}Supplementary Material \\
\vspace{1.0em}
]

\subsection{Influence function}

When training a network by minimizing $\operatorname{D}_\lambda$, the loss function is of the form \Cref{eq:loss_lambda}. To assess the robustness properties, we need to compute the influence function as per \Cref{eq:influence_function}.

First, we compute the first and second derivatives of $\operatorname{D}_\lambda$ with respect to the student outcome probabilities:
\begin{equation}
    \begin{aligned}
        \frac{\partial}{\partial q_j}\operatorname{D}_\lambda\left(p,q\right)&=\frac{\partial}{\partial q_j}\left(\frac{1}{\lambda\left(1+\lambda\right)}\sum_{i=1}^Kp_i\left[\left(\frac{p_i}{q_i}\right)^\lambda-1\right]\right)\\
        &=\frac{1}{\lambda\left(1+\lambda\right)}\sum_{i=1}^Kp_i\frac{\partial}{\partial q_j}\left(\frac{p_i}{q_i}\right)^\lambda\\
        &=\frac{1}{\lambda\left(1+\lambda\right)}\sum_{i=1}^Kp_i\frac{\partial}{\partial q_j}\left(\frac{p_i}{q_i}\right)^\lambda\\
        &=\frac{-1}{1+\lambda}\left(\frac{p_j}{q_j}\right)^{\lambda+1}\\
    \end{aligned}
\end{equation}
Similarly, the second derivatives are
\begin{equation}
    \begin{aligned}
        \frac{\partial^2}{\partial q_j^2}\operatorname{D}_\lambda\left(p,q\right)&=\frac{\partial}{\partial q_j}\left(\frac{-1}{1+\lambda}\left(\frac{p_j}{q_j}\right)^{\lambda+1}\right)\\
        &=\left(\frac{p_j}{q_j}\right)^{1+\lambda}\frac{1}{q_j}
    \end{aligned}
\end{equation}
and $\frac{\partial^2}{q_kq_j}\operatorname{D}_\lambda\left(p,q\right)=0$ if \(j\neq k\). Together, the 
\begin{align}
    \nabla_q\operatorname{D}_\lambda\left(p,q\right)=\frac{-1}{1+\lambda}\left(\left(\frac{p_1}{q_1}\right)^{\lambda+1},\dots,\left(\frac{p_K}{q_K}\right)^{\lambda+1}\right)^T
\end{align}
and the Hessian \(\nabla_q^2\operatorname{D}_\lambda\left(p,q\right)\) as the diagonal matrix
\begin{align}
    \nabla_q^2\operatorname{D}_\lambda\left(p,q\right)=\begin{pmatrix} \left(\frac{p_1}{q_1}\right)^{\lambda+1}q_1^{-1} & 0 & 0 \\ ... & ... & ... \\ 0 & 0 & \left(\frac{p_K}{q_K}\right)^{\lambda+1}q_K^{-1} \end{pmatrix}
\end{align}
Now, the standard formula (see (1) in \cite{koh2017understanding}) for the influence function for the problem of the student probabilities estimation is given by
\begin{equation}
    \begin{aligned}
        \operatorname{IF}\left(\mathcal{L}_\lambda,\left(x,y\right)\right)&=-\left(\nabla_q^2\operatorname{D}_\lambda\left(y,q_\theta\left(\cdot|x\right)\right)\right)^{-1}\nabla_q\operatorname{D}_\lambda\left(y,q_\theta\left(\cdot|x\right)\right)
    \end{aligned}
\end{equation}
By plugging it in, we obtain the \Cref{eq:if_value}:
\begin{align}
    \operatorname{IF}\left(\mathcal{L}_\lambda,\left(x,y\right)\right)=\frac{1}{1+\lambda}q_\theta\left(\cdot|x\right) \nonumber
\end{align}

\subsection{Choice of \texorpdfstring{$\lambda$}{lambda}}

The choice of \(\lambda\) is guided by a combination of principled statistical considerations.  
The power-divergence family \(\operatorname{D}_{\lambda}\) forms a continuous one-parameter bridge between several classical goodness-of-fit statistics.  
In particular,
\[
\lambda = 0 
\;\Rightarrow\; 
D_0(p,q) = \mathrm{KL}(p\|q)
\quad\text{(cross-entropy)},
\]
\[
\lambda = 1 
\;\Rightarrow\; 
D_1(p,q) = \tfrac12 \chi^2(p,q)
\quad\text{(Pearson \(\chi^2\))}.
\]
Thus, by tuning \(\lambda\), we continuously adjust the divergence between likelihood-ratio–type behaviour and Pearson-type behaviour.  
Our choice of \(\lambda\) therefore reflects a balance of these statistical regimes, together with robustness considerations derived from influence function analysis.

The problem of parameter estimation is closely related to hypothesis testing: in the classical setting, the optimal parameter obtained via KL-divergence minimization
\begin{align}
    \hat{\theta}_{\mathrm{KL}}
    \coloneqq 
    \arg\min_{\theta} \sum_{n=1}^N \operatorname{KL}\!\left(y_n,\, q_{\theta}(\cdot \mid x_n)\right),
\end{align}
coincides with the parameter value that is least statistically distinguishable from the true one.  
Equivalently, it maximizes the \(p\)-value of the test
\begin{align}
H_0:\theta_0=\theta 
\qquad\text{vs.}\qquad 
H_1:\theta_0\neq \theta,
\end{align}
whose test statistic is precisely the empirical KL appearing in the minimization above.

This correspondence is fully general for the entire power-divergence family~\cite{read2012goodness}.  
Hence, when parameter estimation is performed by minimizing the power-divergence loss~\eqref{eq:loss_lambda}, the selection of \(\lambda\) can be understood by analyzing the statistical properties of the associated hypothesis test.  
This provides a complementary perspective to the influence-function analysis, and both considerations jointly inform the principled choice of \(\lambda\).

%Drawing on the results of~\cite{read2012goodness}, we summarize the key conclusions:

We base our choice of \(\lambda\) on three observations from the statistical literature (we refer the reader for example to classical textbook by Read and Cressie~\cite{read2012goodness}):
\begin{enumerate}

    \item \textbf{Sensitivity to bump vs.\ dip alternatives.}  
    Smaller values of \(\lambda\) yield tests that are less powerful against \emph{bump alternatives}, that is, if we test a uniform distribution against the alternative \begin{align} H_0: q_{\theta_0}\left(k\right)=\frac{1}{K}, \qquad H_1:q_{\theta_0}=\begin{cases} \frac{1-\delta/\left( K-1\right)}{K}, \quad k\neq K\\ \frac{1+\delta}{K}, \quad k=K \label{eq:alternative} \end{cases} \end{align} for positive \(\delta\), the probability of rejecting $H_0$ when $H_1$ holds is smaller with \(\lambda\) small (or equivalently, adding outliers so that a specific class is over-represented does, the estimation remains robust); vice versa, if \(\delta\) in \Cref{eq:alternative} is negative, i.e. in the case of \textit{dip alternative} (or misspecification due to under-representation of a certain class), the robustness property is stronger for \(\lambda\) larger; if we wish to balance for both cases, we should choose \(\lambda\in[1/3,2/3]\) (see \cite{read2012goodness}, p.82, point c).
    
    \item \textbf{Stability with small expected counts.}  
    When the expected number of samples per class is below \(1\) in given batch, which common for datasets with high number of classes such as CIFAR-100 or ImageNet, the statistic becomes highly unstable unless \(\lambda\in[1/3,\,3/2]\) (see \cite{read2012goodness}, Section~5.4).

    \item \textbf{Need for correction terms.}  
    Only \(\lambda=2/3\) and \(\lambda=1\) yield test statistics that do not require correction terms to maintain adequate test power. Incorporating such corrections would complicate the use of the loss in a training pipeline: the correction factors depend explicitly on the number of classes and the sample size, introducing additional hyper-parameters and undermining one of the main motivations for employing the power-divergence family, namely, obtaining a simple and stable loss function (see~\cite{cressie1984multinomial}, Section~4.1). %  p.~79 in~\cite{read2012goodness};  in   % at mame neco s jinou referenci :)

\end{enumerate}

Combining the three intervals from the above points, we arrive at two reasonable choices -- \(\lambda=2/3\) and \(\lambda=1\). Finally, considering the the influence function of \Cref{eq:influence_function}, we conclude that \(\lambda = 2/3\) is the preferred and most practically robust choice for the power diversity parameter, because it is more efficient in statistical sense.

\subsection{Compatibility with Temperature Scaling}

For applications within some existing training frameworks, we need to pre-process the model logits that enter the divergence loss \(\operatorname{D}_\lambda\) by a temperature scaling. To ensure that the contribution of the divergence term remains comparable across different temperatures, we must understand how the gradients of \(\operatorname{D}_\lambda\) scale with the temperature parameter.  

First, we compute how does the derivatives of \(\operatorname{D}_\lambda\) with respect to the logits of the student scale with the temperature parameter $\tau>0$. We denote the teacher logits as \(u\) and the student logits as \(v\), so that
\begin{align}
    p_i=\frac{e^{u_i}}{\sum_{j=1}e^{u_j}}, \quad q_i=\frac{e^{v_i}}{\sum_{j=1}e^{v_j}}.
\end{align}
We have
\begin{equation}
    \begin{aligned}
        \frac{\partial}{\partial v_j}\operatorname{D}_\lambda\left(p,q\right)&=\frac{\partial}{\partial v_j}\left(\frac{1}{\lambda\left(1+\lambda\right)}\sum_{i=1}^Kp_i\left[\left(\frac{p_i}{q_i}\right)^\lambda-1\right]\right)\\
        &=\frac{-1}{\left(1+\lambda\right)}\sum_{i=1}^K\left(\frac{p_i}{q_i}\right)^{1+\lambda}\frac{\partial q_i}{\partial v_j}\\
        &=\frac{-1}{\left(1+\lambda\right)}\sum_{i=1}^K\left(\frac{p_i}{q_i}\right)^{1+\lambda}q_i\left(\delta_{ij}-q_j\right)\\
    \end{aligned}
\end{equation}
as
\begin{align}
    \frac{\partial q_i}{\partial v_j}=q_i\left(\delta_{ij}-q_j\right).
\end{align}
We further obtain the following.
\begin{equation}
    \begin{aligned}
        &\frac{\partial}{\partial v_j}\operatorname{D}_\lambda\left(p,q\right)=\frac{-1}{\left(1+\lambda\right)}\sum_{i=1}^K\left(\frac{p_i}{q_i}\right)^{1+\lambda}q_i\left(\delta_{ij}-q_j\right)\\
        &=\frac{q_j}{1+\lambda}\left[\sum_{i=1}^Kp_i\left(\frac{p_i}{q_i}\right)^{\lambda}-\left(\frac{p_j}{q_j}\right)^{1+\lambda}\right]\\
        &=\frac{q_j}{1+\lambda}\left[\sum_{i=1}^K\left(\frac{e^{u_i}}{\sum_{k=1}^Ke^{u_k}}\right)^{\lambda+1}\left(\frac{e^{v_i}}{\sum_{k=1}^Ke^{v_k}}\right)^{-\lambda}\right.\\
        &\quad\quad\quad-\left.\left(\frac{e^{u_j}}{e^{v_j}}\right)^{1+\lambda}\left(\frac{\sum_{k=1}^Ke^{v_k}}{\sum_{k=1}^Ke^{u_k}}\right)^{1+\lambda}\right]\\
    \end{aligned}
\end{equation}
We now introduce the temperature scale \(\tau\) and the derivatives become
\begin{equation}
    \begin{aligned}
        &\frac{\partial \operatorname{D}_\lambda\left(p^\tau,q^{\tau}\right)}{\partial v_j}=\\
        &\quad\quad\frac{1}{\tau}\frac{q_j^\tau}{1+\lambda}\left[\sum_{i=1}^K\left(\frac{e^{u_i/\tau}}{\sum_{k=1}^Ke^{u_k/\tau}}\right)^{\lambda+1}\left(\frac{e^{v_i/\tau}}{\sum_{k=1}^Ke^{v_k/\tau}}\right)^{-\lambda}\right.\\
        &\quad\quad\quad-\left.\left(\frac{e^{u_j/\tau}}{e^{v_j/\tau}}\right)^{1+\lambda}\left(\frac{\sum_{k=1}^Ke^{v_k/\tau}}{\sum_{k=1}^Ke^{u_k/\tau}}\right)^{1+\lambda}\right]\\
    \end{aligned}
\end{equation}
(the factor \(1/\tau\) comes from the derivative of \(v_j/\tau\) with respect to \(v_j\)). Now we suppose that both the logits are centered (which has become a standard in KD, \cite{sun2024logit}) and negligible compared to the temperature $\tau$ (temperatures \(\sim 4\) are standard in KD, \cite{hinton2015distilling}), specifically, we assume
\begin{align}
    \left(u_i/\tau\right)^2, \quad \left(v_i/\tau\right)^2, \quad u_iv_i/\tau^2
    \label{eq:negligible}
\end{align}
are negligible. Then using a Taylor polynomial of order one for both nominator and denominator, we obtain an approximation
\begin{equation}
    \begin{aligned}
        \left(\frac{e^{u_i/\tau}}{\sum_{k=1}^Ke^{u_k/\tau}}\right)^\alpha&\sim\frac{1+\alpha u_i/\tau}{N^\alpha}\\
        \left(\frac{e^{v_i/\tau}}{\sum_{k=1}^Ke^{v_k/\tau}}\right)^\alpha&\sim\frac{1+\alpha v_i/\tau}{N^\alpha}
    \end{aligned}
\end{equation}
for any $\alpha>0$. This yields an approximation
\begin{equation}
    \begin{aligned}
        \frac{\partial}{\partial v_j}&\operatorname{D}_\lambda\left(p^\tau,q^{\tau}\right)\\
        &\sim\frac{1}{\tau}\frac{q_j^\tau}{1+\lambda}\left[\sum_{i=1}^K\frac{1+\left(1+\lambda\right)u_i/\tau}{K^{1+\lambda}}\frac{1-\lambda v_i/\tau}{K^{-\lambda}}\right.\\
        &\quad\quad\quad\quad\quad\quad-\left.\frac{1+\left(1+\lambda\right)u_j/\tau}{K^{1+\lambda}}\frac{1-\left(1+\lambda\right)v_j/\tau}{K^{-\left(1+\lambda\right)}}\right]\\
        &=\frac{1}{\tau}\frac{q_j^\tau}{1+\lambda}\left[\frac{1}{K}\sum_{i=1}^K\left(1-\lambda v_i/\tau+\left(1+\lambda\right)u_i/\tau\right.\right.\\
        &\quad\quad\quad\quad\quad\quad\left.\left.-\lambda\left(1+\lambda\right)u_iv_i/\tau^2\right)-1+\left(1+\lambda\right)v_j/\tau\right.\\
        &\quad\quad\quad\quad\quad\quad\quad\quad\left.-\left(1+\lambda\right)u_j/\tau+\left(1+\lambda\right)^2u_jv_j/\tau^2\right]\\
        &=\frac{1}{\tau}\frac{q_j^\tau}{1+\lambda}\left[-\lambda\left(1+\lambda\right)\frac{1}{K}\sum_{k=1}^Ku_iv_i/\tau^2\right.\\
        &\quad\quad\quad\quad\left.+\left(1+\lambda\right)^2u_jv_j/\tau^2+\left(1+\lambda\right)\left(u_j-v_j\right)/\tau\right]
    \end{aligned}
\end{equation}
However, since \(u_iv_i/\tau^2\) is negligible by assumption and we obtain
\begin{align}
    \frac{\partial}{\partial v_j}\operatorname{D}_\lambda\left(p^\tau,q^{\tau}\right)&\sim\frac{q_j^\tau}{\tau^2}\left(u_j-v_j\right)
    \label{eq:temp_scale}
\end{align}
Following the reasoning in Sec.2 of \cite{hinton2015distilling}, we note that the gradients produced by the softened targets scale as \(1/\tau^2\) as can be seen in \Cref{eq:temp_scale}. Consequently, when combining soft and hard targets—as is standard in knowledge distillation, where ground-truth labels (without temperature) are fitted alongside softened outputs, it is important to multiply the loss of soft-targets by \(\tau^2\). This adjustment preserves the relative influence of the hard and soft targets, ensuring that their balance remains stable even as the distillation temperature is varied during hyper-parameter tuning.